\documentclass{article}

\usepackage[preprint]{neurips_2025}

\usepackage[utf8]{inputenc}
\usepackage[T1]{fontenc}

\usepackage{tikz}
\usepackage{float}
\usetikzlibrary{arrows.meta,positioning,shapes.misc}

\usepackage{hyperref}
\usepackage{url}
\usepackage{booktabs}
\usepackage{amsfonts}
\usepackage{nicefrac}
\usepackage{microtype}
\usepackage{xcolor}
\usepackage{graphicx}
\usepackage{amsmath}
\usepackage{tabularx}

\title{Revac: A Social Deduction Reasoning Agent}

\author{%
  Mihir S Arya \\
  Department of Computer Science and Engineering \\
  RV College of Engineering \\
  Bangalore \\
  \texttt{mihirsarya.cy23@rvce.edu.in}
  \And
  Avinash Anish \\
  Department of Information Science and Engineering \\
  RV College of Engineering \\
  Bangalore \\
  \texttt{avinashanish.is23@rvce.edu.in}
  \AND
  Aditya Ranjan \\
  Department of Artificial Intelligence and Machine Learning \\
  RV College of Engineering \\
  Bangalore \\
  \texttt{adityaranjan.ai23@rvce.edu.in}
}

\begin{document}

\maketitle

\begin{abstract}
Social deduction games such as Mafia present a unique AI challenge: players must reason under uncertainty, interpret incomplete and intentionally misleading information, evaluate human-like communication, and make strategic elimination decisions. Unlike deterministic board games, success in Mafia depends not on perfect information or brute-force search, but on inference, memory, and adaptability in the presence of deception. This work presents the design and evaluation of an AI agent, Revac\_8, developed for the Social Deduction track of the MindGames Arena competition, in which the agent achieved first place. The final agent evolved from a simple two-stage reasoning system to a multi-module architecture that integrates memory-based player profiling, social-graph analysis of accusations and defenses, and dynamic tone selection for better communication. The agent's success highlights the critical role of structured memory and adaptive communication in achieving human-level performance in high-stakes social environments. Our agent and code are available at \url{https://github.com/mihiraryaa/mindgames_NeurIPS2025}.
\end{abstract}

\section{Introduction}

Social deduction games, such as Mafia or Werewolf, are complex multi-agent environments requiring sophisticated reasoning and communication skills. Players act under partial observability and asymmetric roles, where success hinges on the ability to infer hidden information, detect deception, and coordinate collective action. This challenge is distinct from traditional board games, demanding an AI capable of not only logical deduction but also social intelligence.

This report documents the development and performance of the Revac agent, which secured first place in the Open Division of the Social Deduction Track of the MindGames Arena's Theory of the Mind Challenge. The agent's development was an iterative process, starting with a basic reasoning loop (Revac) to a multi-faceted architecture of Revac\_8.

The primary objectives of this system were:

\begin{itemize}
\item To perform logically consistent inference grounded in actual game events.
\item To maintain long-term memory of player behavior and relational dynamics.
\item To communicate persuasively while avoiding social suspicion.
\item To remain robust across varied game configurations and player behaviors.
\end{itemize}

We detail the agent's architecture, the development process that led to its final design, the benchmark used to evaluate reasoning quality, and the performance results that validate its approach.

\section{The Game of Secret Mafia}

We evaluate our agent in a turn-based social-deduction setting based on Mafia, where players interact through alternating Night and Day phases under conditions of partial observability and asymmetric roles. A typical setup involves a mix of Village-aligned players (e.g., generic Villagers, Doctor, Detective) and one or more Mafia members who share private knowledge of one another's identities.

During the Night phase, only players with special abilities act privately. For example, the Mafia select a target to eliminate, the Doctor chooses a player to protect, and the Detective investigates a player's alignment---while all actions remain hidden from the rest of the lobby. During the Day phase, the game shifts to open communication: players engage in unrestricted natural-language discussion to reason about observed behaviors, share (or fabricate) evidence, and build consensus before casting votes to eliminate a suspect. This cycle repeats until the Village has eliminated all Mafia members or the Mafia achieve numerical parity with the Village.

\section{Implementation: The Revac\_8 Architecture}

The Revac agent is built upon a modular, multi-stage reasoning architecture designed to decouple complex inference from final action generation, while integrating persistent memory and dynamic communication strategy.

\subsection{Core Two-Stage Reasoning}

The foundational architecture, present even in the initial version (Revac), is a two-stage process:

\begin{enumerate}
\item \textbf{Reviewer Agent}: Takes the current game observation state (chat log, player status, etc.) and generates a detailed, internal chain-of-thought review. This review includes logical deductions, contradiction detection, and a probability assessment of player roles.

\item \textbf{Action Agent}: Takes the original observation state and the detailed review from the Reviewer Agent to formulate the final, natural-language action (a statement, an accusation, or a vote). This separation ensures that the final output is grounded in a deep, structured analysis, minimizing ``hallucinations'' or inconsistent actions.
\end{enumerate}

\subsection{Iterative Evolution of the Agent}

The agent underwent several key iterations, with the most significant advancements occurring between Revac and the final Revac\_8. These improvements focused on enhancing the agent's long-term memory and its social communication capabilities.

\begin{table}[H]
\centering
\caption{Evolution of the Revac agent across versions}
\label{tab:revac-evolution}
\begin{tabular}{lll}
\toprule
Agent Version & Key Architectural Addition & Purpose and Impact \\
\midrule
Revac & Baseline Two-Stage Reasoning & Establishes the core Reviewer-Action loop. \\
      &                              & Lacks memory, leading to short-sighted, \\
      &                              & turn-by-turn decisions. \\
\midrule
Revac2\_1 & Persistent Memory Module & Introduces RevacMemory to track Player \\
          &                          & Profiles and Social Alignment Graph. \\
          &                          & Enables long-term strategic reasoning. \\
\midrule
Revac\_8 & Dynamic Tone Selector & Adds a third stage to select optimal \\
         &                       & communication tone/style. Enhances \\
         &                       & persuasive power and social camouflage. \\
\bottomrule
\end{tabular}
\end{table}

\subsection{The Memory Module: Structured Social Inference}

The Memory Module, introduced in Revac2\_1, is the core innovation enabling Revac to overcome the short-term memory limitations of standard LLM agents. It processes each new observation and updates two key components, with the Social Alignment Graph (SAG) serving as the central mechanism for structured social inference.

\textbf{Player Profiles}: A textual summary of each player's history, including their claims, past votes, and perceived consistency. This provides a long-term behavioral record.

The true novelty lies in the SAG, which transforms raw dialogue into a machine-readable, relational structure.

\subsubsection{The Social Alignment Graph (SAG)}

The Social Alignment Graph is a dynamic, directed graph that models the evolving social landscape of the game. It is a critical departure from simple text-based memory, allowing the agent to perform relational reasoning that is essential for detecting collusion and deception.

\paragraph{Structure and Function}
The SAG represents players as nodes. Edges are weighted and directed, representing specific social interactions extracted from the day-phase discussion. Key edge types include:

\begin{itemize}
\item \textbf{Accusation}: Player A accuses Player B (A $\rightarrow$ B, negative weight).
\item \textbf{Defense/Support}: Player A defends Player B (A $\rightarrow$ B, positive weight).
\item \textbf{Voting Alignment}: Player A votes for Player B (A $\rightarrow$ B, strong negative weight).
\end{itemize}

\paragraph{Novelty and Impact}
The SAG allows the Reviewer Agent to perform graph-based analysis, enabling inferences that are impossible with sequential text alone. For instance:

\begin{enumerate}
\item \textbf{Collusion Detection}: A strong, mutual defense link (A $\leftrightarrow$ B, positive weight) over multiple rounds, especially when one player is under suspicion, strongly suggests a hidden alliance, such as a Mafia partnership.

\item \textbf{Group Pressure Dynamics}: A high in-degree of negative weight (many players accusing one player) signals a group consensus or a potential mislynch, informing the agent's communication strategy (e.g., join the pressure or act as a contrarian).

\item \textbf{Contradiction Grounding}: The graph provides a structured, non-textual anchor for the memory, making the agent's long-term reasoning more robust against the inherent ``forgetting'' or hallucination tendencies of LLMs.
\end{enumerate}

The dynamic nature of the SAG, constantly updated with each turn, provides the agent with a powerful, quantitative tool for social reasoning.

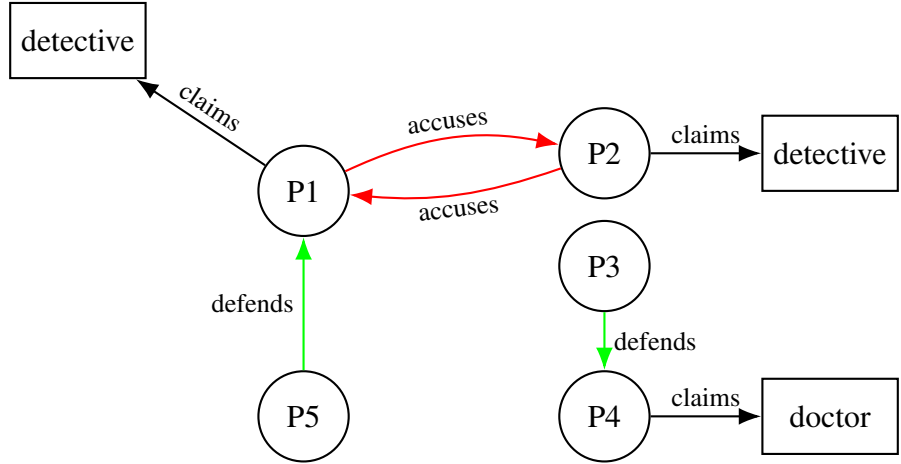
\begin{figure}[H]
    \centering
    \begin{tikzpicture}[node distance=22mm]

    \tikzset{
        person/.style={
            circle,
            draw,
            minimum size=12mm,
            thick,
            font=\large
        },
        rolebox/.style={
            rectangle,
            draw,
            thick,
            minimum width=18mm,
            minimum height=10mm,
            font=\large
        },
        arrow/.style={
            -{Latex[length=3mm]},
            thick
        }
    }

    \node[rolebox] (detL) at (-3, 2) {detective};

    \node[person] (p1) at (0, 0) {P1};
    \node[person] (p2) at (4, 0.5) {P2};

    \node[rolebox] (detR) at (7, 0.5) {detective};

    \node[person] (p5) at (0, -3) {P5};

    \node[person] (p4) at (4, -3) {P4};
    \node[person] (p3) at (4, -1) {P3};

    \node[rolebox] (doc) at (7, -3) {doctor};

    \draw[arrow] (p1) -- node[sloped,above]{claims} (detL);
    \draw[arrow, draw=red] (p1) to[bend left=18] node[sloped,above]{accuses} (p2);
    \draw[arrow, draw=red] (p2) to[bend left=12] node[sloped,below]{accuses} (p1);
    \draw[arrow] (p2) -- node[above]{claims} (detR);
    \draw[arrow, draw=green] (p5) -- node[left]{defends} (p1);
    \draw[arrow, draw=green] (p3) -- node[right]{defends} (p4);
    \draw[arrow] (p4) -- node[above]{claims} (doc);

    \end{tikzpicture}
    \caption{A visual representation of the Social Alignment Graph (SAG) at a critical game state. Nodes represent players, and weighted, directed edges represent social interactions such as accusations (red, negative) and defenses (green, positive).}
    \label{fig:interaction_graph}
\end{figure}

\subsection{Dynamic Tone Selection: The Art of Social Persuasion}

The final and arguably most critical innovation in Revac\_8 is the Dynamic Tone Selector (DTS). In a social deduction game, communication is not merely a way for conveying strategy; it is the strategy itself. A perfectly logical deduction, if delivered in a dry, robotic, or socially inappropriate manner, will fail to persuade human players or agents and may even draw suspicion. The DTS addresses this by ensuring the agent's output is not only logically sound but also socially and rhetorically effective.

The DTS operates as a third stage in the pipeline, taking the Reviewer Agent's deep analysis and the current game state to select an optimal communication style. This selection is dynamic and context-aware, moving beyond a single persona to a collection of social behaviors.

\begin{table}[htbp]
\centering
\caption{Key tones and strategic use cases}
\label{tab:tones}
\begin{tabular}{p{3.5cm}p{4cm}p{5.5cm}}
\toprule
Tone/Style & Strategic Goal & Example Use Case \\
\midrule
Aggressive/Pressuring & Force a suspect to make a mistake or counter-claim. & Used by a Villager when SAG shows a player is weakly defended but highly suspicious. \\
\midrule
Withdrawing/Passive & Deflect suspicion from the agent itself. & Used by Mafia after a failed lie, or by Villager to observe debate without becoming target. \\
\midrule
Logically Anchoring & Establish consensus or redirect discussion to confirmed fact. & Used by Villager to summarize evidence and propose clear, rational vote. \\
\midrule
Contrarian/Skeptical & Break a false consensus or test strength of claim. & Used when SAG shows strong but potentially manufactured group pressure. \\
\bottomrule
\end{tabular}
\end{table}

This module is what allows Revac\_8 to operate convincingly in these environments, where human-like social intelligence is very important. It transforms the agent from a logical deduction machine into a persuasive social actor, capable of both subtle deception and powerful, conviction-based rhetoric.

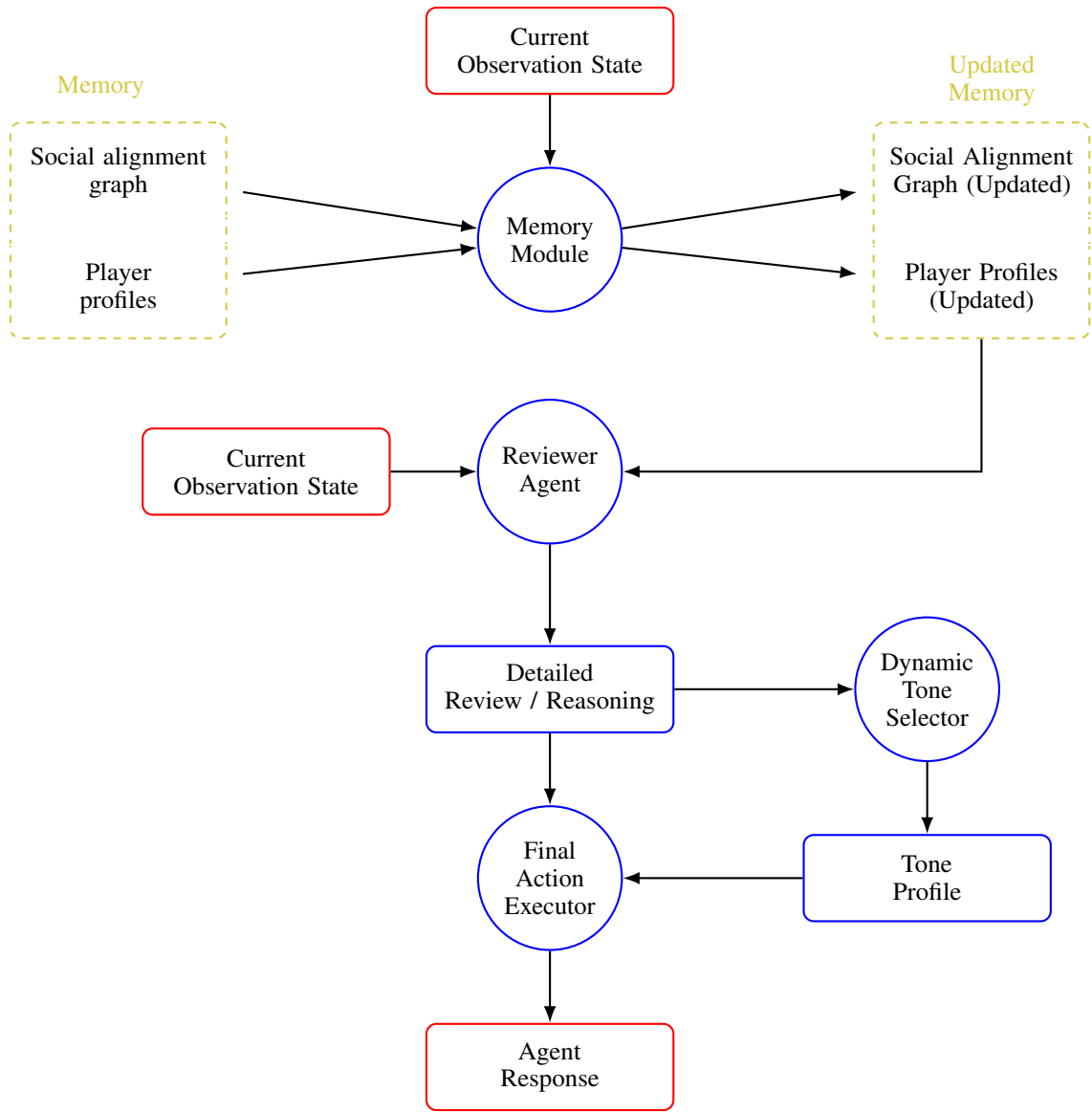
\begin{figure}[H]
\centering
\begin{tikzpicture}[
    node distance=20mm,
    box/.style args={#1}{
        rectangle, rounded corners,
        draw=#1, thick,
        text width=32mm,
        align=center,
        minimum height=12mm
    },
    circ/.style args={#1}{
        circle,
        draw=#1,
        thick,
        minimum size=20mm,
        align=center
    },
    dashedbox/.style args={#1}{
        rectangle, rounded corners,
        draw=#1,
        dashed,
        thick,
        inner sep=5mm
    },
    arrow/.style={-Latex, thick}
]

\node[dashedbox=yellow!80!black, minimum width=3cm, minimum height=3cm] (memL) at (-6,1.5) {};
\node[box=white] (sgraph) at (-6,2.3) {Social alignment\\graph};
\node[box=white] (pprofile) at (-6,0.7) {Player\\profiles};

\node[anchor=south east, align=center, text=yellow!80!black, xshift=20mm, yshift=2mm]
    at (memL.north west) {Memory};

\node[dashedbox=yellow!80!black, minimum width=3cm, minimum height=3cm] (memR) at (6,1.5) {};
\node[box=white] (sgraphU) at (6,2.3) {Social Alignment\\Graph (Updated)};
\node[box=white] (pprofileU) at (6,0.7) {Player Profiles\\(Updated)};

\node[anchor=south west, align=center, text=yellow!80!black, xshift=-21mm, yshift=1mm]
    at (memR.north east) {Updated\\Memory};

\node[box=red] (obs) at (0,4) {Current\\Observation State};
\node[circ=blue, below=10mm of obs] (reviewer) {Memory\\Module};

\node[circ=blue, below=12mm of reviewer] (mid1) {Reviewer\\Agent};
\node[box=red, left=12mm of mid1] (mid2) {Current\\Observation State};

\node[box=blue, below=14mm of mid1] (reason) {Detailed\\Review / Reasoning};
\node[circ=blue, right=25mm of reason] (toneSel) {Dynamic\\Tone\\Selector};
\node[box=blue, below=10mm of toneSel] (toneProf) {Tone\\Profile};
\node[circ=blue, below=10mm of reason] (executor) {Final\\Action\\Executor};
\node[box=red, below=10mm of executor] (resp) {Agent\\Response};

\draw[arrow] (obs) -- (reviewer);

\draw[arrow] (sgraph) -- (reviewer);
\draw[arrow] (pprofile) -- (reviewer);

\draw[arrow] (reviewer) -- (sgraphU);
\draw[arrow] (reviewer) -- (pprofileU);

\draw[arrow] (memR) |- (mid1);
\draw[arrow] (mid2) -- (mid1);
\draw[arrow] (mid1) -- (reason);

\draw[arrow] (reason) -- (toneSel);
\draw[arrow] (toneSel) -- (toneProf);
\draw[arrow] (toneProf) -- (executor);

\draw[arrow] (reason) -- (executor);
\draw[arrow] (executor) -- (resp);

\end{tikzpicture}
\caption{Architecture of Revac\_8}
\label{fig:memory-architecture}
\end{figure}

\section{Evaluation}

The Revac\_8 agent was evaluated using a two-fold approach, reflecting the dual challenge of the Social Mafia game: strategic reasoning (internal deduction) and effective communication (external action).

\subsection{Competition Results}

The Revac agent achieved first place in the Open Division of the Social Deduction Track of Mindgames NeurIPS 2025.

\begin{table}[htbp]
\centering
\begin{tabular}{lr}
\toprule
\textbf{Agent Name} & \textbf{TrueSkill Rating} \\
\midrule
Revac\_8 & 13.9 \\
Fractal\_SecretMafia\_Agent\_round2\_v25 & 7.8 \\
Fractal\_SecretMafia\_Agent\_round2\_v14u & 4.7 \\
\bottomrule
\end{tabular}
\vspace{6pt}
\caption{TrueSkill Ratings of the Top 3 Performing Agents}
\end{table}

\subsection{Two-Fold Evaluation: Strategy and Communication}

\subsubsection{Strategy Evaluation on Benchmark}

We utilize a specialized benchmark to quantitatively assess the agent's internal strategic reasoning. The benchmark contains 13 curated test cases that feature challenging scenarios: multiple conflicting role claims, player hallucinations, no-kill nights, and strategic deception patterns. Each case includes the observation state, ground truth roles, and scenario explanations. This output is scored using two metrics:

\begin{enumerate}
\item \textbf{Metric A: Role Identification Accuracy (Objective)}: Measures the correctness of the agent's predicted roles against the ground truth. The score is calculated as the proportion of correctly identified player roles:
\[
\text{Score} = \frac{\text{correct role predictions}}{\text{total players}}.
\]
This metric ranges from 0.0 (no correct predictions) to 1.0 (all roles correctly identified), providing a direct measure of the agent's deductive accuracy.

\item \textbf{Metric B: Reasoning Quality (LLM-Judged)}: An external LLM judge evaluates the agent's detailed explanation on a 0--5 scale, assessing the logical soundness, use of game evidence, contradiction detection, and avoidance of hallucination. The raw score (0--5) is normalized to 0.0--1.0 for the final scoring.
\end{enumerate}

The combined score is calculated as a weighted average:
\begin{equation}
\text{Final Score} = (0.5 \times \text{Metric A}) + (0.5 \times \text{Metric B}_{\text{Normalized}})
\end{equation}

\begin{table}[htbp]
\centering
\caption{Benchmark Evaluation Results with different models}
\label{table:accuracy}

\resizebox{0.8\textwidth}{!}{%
\renewcommand{\arraystretch}{1.2}
\begin{tabular}{lccc}
\toprule
\textbf{Model} & \textbf{Metric A} & \textbf{Metric B (Norm)} & \textbf{Final Score} \\
\midrule
\multicolumn{4}{c}{\textbf{Revac}} \\
\midrule
gpt-5(low) & 0.74 & 0.52 & 0.63 \\
gpt-5-mini & 0.62 & 0.71 & 0.66 \\
kimi-k2-instruct & 0.78 & 0.40 & 0.59 \\
\midrule
\multicolumn{4}{c}{\textbf{Revac2.1}} \\
\midrule
gpt-5(low) & 0.78 & 0.70 & 0.74 \\
gpt-5-mini & 0.74 & 0.58 & 0.66 \\
kimi-k2-instruct & 0.82 & 0.53 & 0.68 \\
\midrule
\multicolumn{4}{c}{\textbf{Revac\_8}} \\
\midrule
gpt-5(low) & 0.78 & 0.66 & 0.72 \\
gpt-5-mini & 0.89 & 0.70 & 0.80 \\
kimi-k2-instruct & 0.80 & 0.50 & 0.65 \\
\bottomrule
\end{tabular}%
}
\end{table}

This two-metric approach ensures that the agent is rewarded not just for guessing the correct roles, but for demonstrating a robust, human-like chain of thought, which is the true measure of its strategic capability. Since the benchmark only contains 13 examples, it does not fully capture the agent's reasoning ability. Instead, it gives a general idea of how the agent thinks through complicated situations. The best demonstration of its reasoning comes from real gameplay, where it adapts to different players and unpredictable discussions.

\subsubsection{Qualitative Analysis of Strategic Reasoning}

Because reasoning in Mafia is very subjective and dynamic, we illustrate the strategic thinking of the agent through examples taken from actual matches. These cases show how the Revac\_8 agent tracks claims, spots inconsistencies, and forms justified conclusions in context.

\begin{table}[htbp]
\centering
\begin{tabular}{|p{4cm}|p{5cm}|p{5cm}|}
\hline
\textbf{Game State Summary} & \textbf{Agent's Reasoning (Revac\_8)} & \textbf{Ground Truth Analysis} \\ \hline

P0 claims Detective, checked P3 as Mafia. P3 counter-claims Detective, checked P0 as Mafia. P4 claims Doctor, protected P3. P5 counter-claims Doctor, saved self. &
Revac8 identifies the core contradiction (two Detectives, two Doctors). It uses the Social Alignment Graph to note P4's defense of P3 (``I protected P3'') as suspiciously convenient, especially since P5's self-save plausibly explains the ``no kill'' night log. Revac\_8 concludes the most likely scenario is P3 is Mafia and P4 is their partner providing cover. &
P3 Mafia, P0 real Detective; P4 Mafia partner, P5 real Doctor. Revac\_8 successfully navigated the web of lies by prioritizing the analysis of who benefits from the claims and which claims are most consistent with the game log. \\ \hline
\end{tabular}
\vspace{6pt}
\caption{Case 1: Conflicting Claims and Strategic Defense}
\end{table}

\begin{table}[htbp]
\centering
\begin{tabular}{|p{4cm}|p{5cm}|p{5cm}|}
\hline
\textbf{Game State Summary} & \textbf{Agent's Reasoning (Revac\_8)} & \textbf{Ground Truth Analysis} \\ \hline

P1 claims Villager but talks about ``investigating P0'' (who is dead). P2 makes an unclear claim. P4 pushes for a structured mass-claim. P3 remains silent. P2 votes for P5 (who is quiet). &
Revac8's memory flags P1's claim as inconsistent (``investigating P0'' is impossible for a Villager and illogical for a Detective). It notes P3's silence as suspicious. The Social Alignment Graph notes P2's early vote on P5 as ``opportunistic/dangerous'' at LyLo, suggesting a possible attempt to steer the vote. Revac\_8 focuses on forcing full claims from P2 and P3 before voting. &
P3 is Mafia. Revac\_8's focus on claim consistency and strategic voting behavior (P2's opportunistic vote) demonstrates a deeper social understanding than a purely logical agent. \\ \hline
\end{tabular}
\vspace{6pt}
\caption{Case 2: Inconsistent Claims and Opportunistic Voting}
\end{table}

\begin{table}[htbp]
\centering
\begin{tabular}{|p{4.5cm}|p{5cm}|p{5cm}|}
\hline
\textbf{Game State Summary} & \textbf{Agent's Reasoning (Revac\_8)} & \textbf{Ground Truth Analysis} \\ \hline

P1 full-claims Detective, checked P2 as Mafia on N0. P5 claims Doctor, protected P2 on N1. P2 counter-claims Mafia, says P1 is lying. P0 makes host-like comments. &
Revac\_8 immediately recognizes the high-stakes LyLo scenario. It prioritizes the confirmed information: P1's claim on P2 is the strongest lead. It questions P5's protection of P2 (``why did you protect 2 specifically last night?''). It also flags P0's ``host-like'' comments as suspicious TMI (Too Much Information). Revac8 supports lynching P2 but demands clarification on P5 and P0 before hammering the vote. &
P2 is Mafia. Revac8's ability to process the full-claim, identify the most critical piece of evidence (P1's check), and simultaneously question the supporting/suspicious actions of others (P5's protection, P0's TMI) showcases its multi-faceted reasoning. \\ \hline
\end{tabular}
\vspace{6pt}
\caption{Case 3: Full Claim and Counter-Claim at LyLo}
\end{table}

\subsubsection{Qualitative Analysis of Communication (Dynamic Tone)}

Apart from the internal strategy, the success of Revac\_8 is majorly dependent on its ability to execute its strategy through persuasive communication. The Dynamic Tone Selector (DTS) makes sure that the agent's output is not dry and logical, but tailored to the social context of the game.

\begin{table}[H]
\centering
\caption{Examples of dynamic tone selection}
\label{tab:tone-examples}
\begin{tabular}{p{1.5cm} p{1.4cm} p{5cm} p{5cm}}
\toprule
Strategic Goal & Tone Selected (DTS) & Agent's Output (Example) & Impact \\
\midrule
Pressure a Suspect & Aggressive / Demanding & ``Player 3, your silence is deafening. We are at LyLo. If you are town, you have one chance to full-claim your role and N1 action now. Otherwise, you are my lynch vote.'' & Forces an immediate, high-stakes response, often leading to a slip-up or a forced claim, which is then analyzed by the Reviewer. \\
\midrule
Deflect Suspicion (Mafia) & Withdrawing / Confused & ``Okay, okay, everyone calm down. I was just trying to figure out what was going on. I'm just a normal villager guys, no special powers. I agree Player 5 is super sus, but now Player 1 throwing shade at me?'' & Mimics a panicked, innocent player, effectively diverting the focus from the agent's lack of a strong claim and onto the general chaos. \\
\midrule
Establish Authority (Villager) & Logically Anchoring & ``Let's reset. We have one confirmed fact: Player 1's check on Player 2. All other claims are secondary until we address this. We lynch 2 today, or we lose. Any argument against this is a Mafia distraction.'' & Uses a firm, structured tone to cut through noise and anchor the town's collective reasoning to the most critical piece of evidence, minimizing mislynches. \\
\bottomrule
\end{tabular}
\end{table}

\section{Conclusion}

The success of the Revac\_8 agent in the MindGames Arena Mafia competition is a testament to the importance of integrating structured memory and dynamic communication into large language model based agents. The evolution from a stateless agent (Revac) to the memory-equipped, socially-aware Revac\_8 shows that achieving human-level performance in social deduction requires moving beyond simple prompt engineering to a multi-module, architectural approach.

\subsection{Lessons Learned}

The key takeaway is that in a social deduction environment, communication is a strategic action. The Dynamic Tone Selector proved invaluable, allowing the agent to shift its persona from a logical investigator to a persuasive ally or a deceptive opponent, depending on the game state and its current role. Furthermore, the persistent memory, particularly the Social Alignment Graph, enabled the agent to detect subtle, long-term patterns of collusion and contradiction that are not evident in a turn-by-turn reasoning agent.

\section{Limitations}

A key limitation of our approach is that the current observation state analysis technique is dependent on the task. The structure of the Social Alignment Graph and the Player Profiles are manually specified to extract features relevant to Mafia (e.g., accusations, defenses, claims). While this structure applies to a wide range of social deduction games, an interesting future direction would be to allow agents to identify which aspects of the scene are relevant to a specific task instead of manually specifying the feature extraction. This would move the system closer to a truly general-purpose social reasoning AI.

Another limitation concerns the benchmark used to quantify reasoning ability. Because it contains only 13 curated cases, it cannot fully represent the diversity of real-game situations or the agent's long-term reasoning behavior. The benchmark gives a helpful snapshot of reasoning quality, but not a complete or statistically robust measure of performance. The agent's reasoning is much better demonstrated in live matches using a qualitative analysis, where it must adapt to evolving discussions, unpredictable behaviors, and deception.

\end{document}